\documentclass[sigconf]{acmart}

\AtBeginDocument{%
  \providecommand\BibTeX{{%
    \normalfont B\kern-0.5em{\scshape i\kern-0.25em b}\kern-0.8em\TeX}}}

\usepackage[misc,geometry]{ifsym}

\begin{document}
\pagestyle{plain}
%%
%% The "title" command has an optional parameter,
%% allowing the author to define a "short title" to be used in page headers.
\title{Industry and Academic Research in Computer Vision}

%%
%% The "author" command and its associated commands are used to define
%% the authors and their affiliations.
%% Of note is the shared affiliation of the first two authors, and the
%% "authornote" and "authornotemark" commands
%% used to denote shared contribution to the research.
\author{Iuliia Kotseruba, Manos Papagelis, John K. Tsotsos}
\email{yulia_k,papaggel,tsotsos@eecs.yorku.ca}
\affiliation{%
  \institution{York University}
  \city{Toronto}
  \state{ON}
}

\settopmatter{printacmref=false}
\setcopyright{none}
\renewcommand\footnotetextcopyrightpermission[1]{}

%%
%% By default, the full list of authors will be used in the page
%% headers. Often, this list is too long, and will overlap
%% other information printed in the page headers. This command allows
%% the author to define a more concise list
%% of authors' names for this purpose.
\renewcommand{\shortauthors}{Kotseruba,Papagelis,Tsotsos}

%%
%% The abstract is a short summary of the work to be presented in the
%% article.
\begin{abstract}
This work aims to study the dynamic between research in the industry and academia in computer vision. The results are demonstrated on a set of top-5 vision conferences that are representative of the field. Since data for such analysis was not readily available, significant effort was spent on gathering and processing meta-data from the original publications. 

First, this study quantifies the share of industry-sponsored research. Specifically, it shows that the proportion of papers published by industry-affiliated researchers is increasing and that more academics join companies or collaborate with them. Next, the possible impact of industry presence is further explored, namely in the distribution of research topics and citation patterns. The results indicate that the distribution of the research topics is similar in industry and academic papers. However, there is a strong preference towards citing industry papers. Finally, possible reasons for citation bias, such as code availability and influence, are investigated.
\end{abstract}

%%
%% The code below is generated by the tool at http://dl.acm.org/ccs.cfm.
%% Please copy and paste the code instead of the example below.
%%
%\begin{CCSXML}
%<ccs2012>
% <concept>
%  <concept_id>10010520.10010553.10010562</concept_id>
%  <concept_desc>Computer systems organization~Embedded systems</concept_desc>
%  <concept_significance>500</concept_significance>
% </concept>
% <concept>
%  <concept_id>10010520.10010575.10010755</concept_id>
%  <concept_desc>Computer systems organization~Redundancy</concept_desc>
%  <concept_significance>300</concept_significance>
% </concept>
% <concept>
%  <concept_id>10010520.10010553.10010554</concept_id>
%  <concept_desc>Computer systems organization~Robotics</concept_desc>
%  <concept_significance>100</concept_significance>
% </concept>
% <concept>
%  <concept_id>10003033.10003083.10003095</concept_id>
%  <concept_desc>Networks~Network reliability</concept_desc>
%  <concept_significance>100</concept_significance>
% </concept>
%</ccs2012>
%\end{CCSXML}

%\ccsdesc[500]{Computer systems organization~Embedded systems}
%\ccsdesc[300]{Computer systems organization~Redundancy}
%\ccsdesc{Computer systems organization~Robotics}
%\ccsdesc[100]{Networks~Network reliability}

%%
%% Keywords. The author(s) should pick words that accurately describe
%% the work being presented. Separate the keywords with commas.
\keywords{computer vision, industry, academia, citation network, affiliation network, bibliographic study}

%% A "teaser" image appears between the author and affiliation
%% information and the body of the document, and typically spans the
%% page.
%\begin{teaserfigure}
%  \includegraphics[width=\textwidth]{sampleteaser}
%  \caption{Seattle Mariners at Spring Training, 2010.}
%  \Description{Enjoying the baseball game from the third-base
%  seats. Ichiro Suzuki preparing to bat.}
%  \label{fig:teaser}
%\end{teaserfigure}

%%
%% This command processes the author and affiliation and title
%% information and builds the first part of the formatted document.
\maketitle

\section{Introduction}
\label{motivation}

Industry-sponsored and academic research have coexisted and interacted in all areas of science despite the fact that there are significant differences in the way research is conducted in both. Research in companies is often motivated by short-term goals aligned with business needs and specific product features rather than pure academic inquiry. The work-flows also differ. For instance, in machine learning research, academics aim to develop new models or improve existing models to establish new state-of-the-art. In comparison, industry research often works backward, i.e. it starts with fixed performance requirements and looks for a specific model or approach that can satisfy them and scale up easily \cite{rothe_2018}. As a result of different objectives, some industry leaders admit that the next great idea is more likely to come from academia \cite{sahuguet_2016}. At the same time, the industry has vastly more resources and data not available in the academia.

I will look at the dynamic between industry and academia in the field of computer vision, where research activity has been increasing in the past decade. Much of this growth is attributed to the success of deep learning which, in turn, was enabled by the availability of computational resources and large-scale data. Given the enormous value of computer vision applications, many companies invest in R\&D in this area. As a result, industry-sponsored researchers became a prominent part of the computer vision community and contributed to all the major conferences and journals in the field. In addition, all top computer vision conferences feature industry expos which serve as opportunities for networking and recruitment.

Although the industry's growing presence is hard to ignore, its impact on computer vision research is largely unknown due to the lack of relevant data and formal studies. Therefore, the goal of this study is two-fold: to quantify the share of industry-sponsored research in the field of computer vision and to understand whether industry presence has a measurable effect on the way the field is developing.

%To address the first problem we can ask the folowing questions. What is the proportion of the industry-sponsored research within computer vision community and how it changed over the years? What are the companies that publish the most and in what areas of computer vision?  Given author affiliation data over multiple years it will be possible to study the movement of researchers between academia and industry. Provided there are some noticeable trends, is it possible to determine whether researchers tend to publish/collaborate more or less once they move to the industry? 

%
%To minimize subjective judgment we decided to look at the evolution of the research topics over the past decade and determine whether the presence of the industry can be attributed to these changes. For instance, are there areas with higher concentration of industry-sponsored research and does this in turn attract more academic research? Are there areas in computer vision that became depleted or overrepresented?

\section{Previous work}
\label{previous_work}
The scale of the industry involvement and its dynamics can be glimpsed from several informal studies that analyzed papers accepted in the top machine learning conferences, such as International Conference on Machine Learning (ICML) and Conference on Neural Information Processing Systems (NeurIPS). Given the significant overlap between vision and machine learning communities, conclusions from these studies should apply to the computer vision field as well. One study found that the number of papers in ICML where at least one author was affiliated with the industry increased from 20-25\% in 2017 \cite{karpathy_2017} to 45\% in 2018 \cite{guliani_2018}. Publications by researchers from Google's DeepMind alone doubled from 6\% in ICML'17 \cite{karpathy_2017} to 13\% in ICML'18 \cite{guliani_2018}. A different analysis of ICML and NeurIPS in 2019 \cite{chuvpilo_2019} indicates the industry presence in 22.2\%  of papers (also confirmed by \cite{andreasdoerr_2019}). The author concluded that despite significant industry presence, there were no signs of monopolization of AI research by the industry in 2019. Another study of NeurIPS in 2014-2018 and ICML in 2017 and 2018 found that industry papers were among the most cited and that many influential authors were collaborating with companies \cite{raghu_2019}. It should be noted that while these studies detect similar trends, they rely on noisy data, which lacks temporal dimension. Therefore, a more rigorous analysis is needed before any conclusions can be made.

\section{Dataset}
In 2012 Kryzhevsky et al. presented AlexNet, which demonstrated the advantages of convolutional neural networks (CNNs) over other machine learning approaches for large-scale image classification tasks \cite{krizhevsky2012imagenet}. Shortly after, CNNs and deep learning spread to many other areas in computer vision and beyond. For this study, to capture trends due to the (re-)introduction of this machine learning paradigm, data will be mined from publications accepted to major computer vision conferences from 2010 to 2019. The following conferences were selected: Conference on Computer Vision and Pattern Recognition (CVPR), International/European/Asian Conference on Computer Vision (ICCV, ECCV, ACCV), and British Machine Vision Conference (BMVC). There are a total of 14,686 papers in the dataset.

\section{Method}

\subsection{Meta-data extraction}

To analyze the presence and impact of the industry in the computer vision community, it is necessary to have the following information: \textit{authors} and their \textit{affiliations} (with temporal correspondence), \textit{affiliation type} (industry or academia), \textit{papers}  and corresponding \textit{year}, \textit{venue}, \textit{titles} \textit{references}, \textit{abstracts}, and \textit{topics}.

Affiliation data for each author with temporal correspondence is not readily available (as pointed out in the previous studies \cite{karpathy_2017,guliani_2018,raghu_2019}). While most computer vision conference papers are available in open access, only the author names and paper titles can be easily obtained from the conference proceedings and/or conference web pages. Alternative sources such as Google Scholar, \texttt{dblp}, and Crossref do not provide the required information. For example, Google Scholar provides only the current affiliation information, and in both \texttt{dblp} and Crossref, affiliations are missing for most authors.

The authors of previous studies relied primarily on manual clean-up of the data. However, given the large scope of the present study and time constraints, purely manual data extraction was infeasible, therefore heuristics and specialized software were used to pre-process the data and reduce manual clean-up. The affiliation extraction pipeline is described below.

\textbf{Downloading PDFs of the papers.}  We created multiple \texttt{python} scripts to download paper pdfs from open-access proceedings. These include BMVC, ICCV (since 2013), and CVPR (since 2013). The publisher of ACCV and ECCV provides institutional access to the full proceedings, which can be downloaded as a single pdf and split into individual papers (e.g. using \texttt{pdfsam}\footnote{\url{https://pdfsam.org/}} tool). The papers from CVPR and ICCV prior to 2013 could not be downloaded from the publisher's website since it prohibits scraping and the full proceedings in a single pdf were not available. We gathered a list of publications from the conference programs and used scripts to query Google Scholar to find pdfs of these papers from alternative sources online.

\textbf{Parsing paper headers.} To extract header meta-data, specifically, title, abstract, authors' names, and affiliations, we used GROBID \cite{GROBID}, an actively developed library that applies a combination of machine learning and rule-based logic to parse pdf files. We ran GROBID on the entire dataset of 14,686 papers to generate TEI\footnote{\url{https://tei-c.org/}} compliant XML for each paper. To make XML files more human-readable, we stripped the TEI tags and removed extra information such as department titles, authors' emails, and affiliation addresses. Out of the entire corpus, approximately $1\%$ of pdfs could not be parsed automatically due to various reasons, such as missing title, author and/or affiliation information, wrong file format, corrupted pdf or pdf submitted as an image. These were processed manually.

Unfortunately, such automatic data extraction is noisy. GROBID had trouble parsing some affiliation names, especially if they contained human-name-like words, e.g. ``University of Michigan at Ann Arbor'' or ``Virginia Tech'' (these were treated as authors). Some other word combinations were often parsed as author names, e.g. ``Microsoft Research'', ``Remote Sensing'', ``ETH Zurich''. Different types of errors occurred depending on the specific paper templates. For example, the BMVC template contains a header that GROBIT often erroneously appended to the title of the paper. In the ECCV template, corresponding author's names were not parsed correctly in the presence of the envelope icon (\Letter). GROBID also incorrectly parsed long names with more than two components: a single long name often was split erroneously into two, increasing the author count. Often, parts of the affiliation string were appended to the authors' last names or the title of the paper. Some parsing issues were also due to the errors introduced by the authors of the papers, e.g. misspelled or missing affiliation names, modified paper templates, etc.

To make sure that parsed meta-data was accurate, we compared parsed titles and the number of authors against the data extracted from conference programs and checked for any missing author affiliations. In addition, we used Stanford Named Entity Recognizer (NER) implemented in \texttt{nltk} \cite{bird2009natural} to ensure that authors' names are human names and not organizations, locations, or random words. Parsed data for more than $50\%$ of papers failed one or more of these checks and had to be corrected manually.

\textbf{Resolving affiliation names.} Since templates do not provide the standard way of specifying affiliations, there is a significant variation in the spelling of the same affiliation names. There are several recurring themes across the dataset: 1) inconsistent use of the official and colloquial institution name (e.g. ``UCLA'' vs ``University of California, Los Angeles'',  ``Caltech'' vs ``California Institute of Technology''), 2) inconsistent level of affiliation details (e.g. only the university name is provided or the full hierarchy including lab, department, institute, funding agency), 3) multiple affiliation levels specified due to the convoluted educational systems in some countries (e.g. multi-layer structure of government research agencies and universities in China and France), 4) incorrectly spelled or parsed affiliations and 5) inconsistent format for specifying multiple affiliations for multiple authors (e.g. missing or incorrect superscripts, superscripts before or after the authors' names, use of non-ASCII characters which occasionally break the parser).

A combination of all these factors made it very challenging to automatically standardize affiliation names across the dataset, therefore this step was performed manually. A list of all parsed affiliations was clustered by similarity using Levenshtein distance \cite{levenshtein1966binary}. Then we manually corrected misspelled affiliation names, expanded all abbreviations, excluded lab and department names so that only affiliations at the level of the institution (e.g. university, academy or research center) remained. Likewise, for companies, we excluded department and area names since they are inconsistently provided in the papers (e.g. ``Microsoft Research Asia'',  ``Microsoft Research Europe'', and simply ``Microsoft Corporation'' in some papers).

For this study, we also needed to label affiliations as belonging to the industry or academia. We labeled public and private educational institutions, government-sponsored research centers (e.g. U.S. Army Laboratory), and other non-profit organizations as \textit{academia}. All entities that engage in any form of for-profit commercial activities (including start-ups) were labeled as \textit{industry}. There is a growing number of research institutions that do not conduct business but are operated and funded by companies (e.g. DOMO Academy Alibaba, Toyota Research Institute in Chicago, Google's DeepMind, and IBM Thomas J. Watson Research Center). These are also labeled as belonging to the industry.

\textbf{Mining reference data.} We parsed the bibliographies for each paper using GROBID and stripped TEI tags as described earlier. To build a citation network, we chose Digital Object Identifier (DOI)\footnote{\url{https://www.doi.org/}} provided via Crossref REST API\footnote{\url{https://www.crossref.org/services/metadata-delivery/rest-api/}} to serve as a unique identifier for the referenced papers. 

We started by forming a query to Crossref API by concatenating the family name of the first author (if available), the title of the paper, venue and year of publication. We then computed the Levenshtein distance between the titles in the top-5 results returned by the Crossref and the title of the reference. We selected the top match if it was above a set threshold and used the corresponding DOI as the id for that paper. We also collected the authors' names, venue, and year of publication from Crossref for consistency across the dataset. 

We encountered numerous issues with the Crossref database. For example, the names of the authors are often not separated into given and family, and the formats of the venues are inconsistent (e.g. ``CVPR, 2013'' vs ``2013 IEEE Conference on Computer Vision and Pattern Recognition''). Furthermore, since it is a public API, it imposes a limit on the number of queries per minute. Given that there are close to 700K references in the entire dataset, the process took over a week to complete. 

Nearly 10\% of all references are missing DOIs. We selected a random set of references to assess the reasons for missing DOI information. One of the causes is the presence of errors in the parsed titles and incorrectly entered titles in the Crossref database. Besides, some references do not have DOIs assigned. These include references to software, non-academic publications (e.g. technical reports, personal blogs, newspapers, etc.), and pre-prints (e.g. arXiv).

\begin{table}[]
\centering
\resizebox{1\columnwidth}{!}{%
\begin{tabular}{l|c|c|c|c}
                       & Academia & Academia/Industry & Industry & Total   \\ \hline
Number of papers       & 10,739   & 3,399             & 548      & 14,686  \\
Number of authors      & 14,400   & 2,140             & 1,865     & 18,405  \\
Number of citations    & 108,867  & 39,772            & 12,875   & 161,514 \\
Number of affiliations & 1,279    & -                 & 473      & 1,752   \\
Number of code links   & 1,793    & 688               & 69       & 2,550   \\ \hline
\end{tabular}
}%
\caption{Dataset statistics}
\vspace{-3.5em}
\label{data_stats}
\end{table}

\textbf{Finding available code for papers}. Code and data sharing is becoming more widespread in the computer vision community. Although it does not guarantee reproducibility of the results, it makes benchmarking of the models easier, allows others to reuse code in their research, and also appears to boost citations \cite{vandewalle2019code}. Projects such as \url{paperswithcode.com} provide links to code associated with papers and aggregate the reported results. Their database of papers with code is available for download. However, their database is incomplete, with no code links for papers published before 2013. This database indexes code for arXiv pre-prints, some of which later appear in conference proceedings. Because the paper titles may change between the revisions, it is difficult to automatically establish correspondence between the arXiv and conference publications.

In addition to the \texttt{paperswithcode} database, a combination of regular expressions was used to find code links provided in the text of the paper, but only a small portion of papers contain links. Since 2014 the vast majority of code is published on \url{github.com} and can be found via GitHub search API using paper title and conference as a query. Pre-2014 code is much more challenging to find since most of it is hosted on the personal web pages of the researchers or institutional web pages. These links were discovered via manual search and scripts using DuckDuckGo Instant Answer API. A total of 2,550 active links to code/data were found.

\subsection{Data analysis}

\textbf{Dataset statistics.} Table \ref{data_stats} lists the overall statistics of the dataset used for the analysis. Processed and cleaned-up data was stored in the PostgreSQL relational database with the following tables: \texttt{papers} - stores venue, year, paper ids and titles, \texttt{authors} - stores author id, authors first and last names, paper id and corresponding affiliation id(s), and \texttt{affiliations} table stores affiliation id, name and type (industry or academia).

\begin{figure}[t!]
  \centering
  \includegraphics[width=0.85\linewidth]{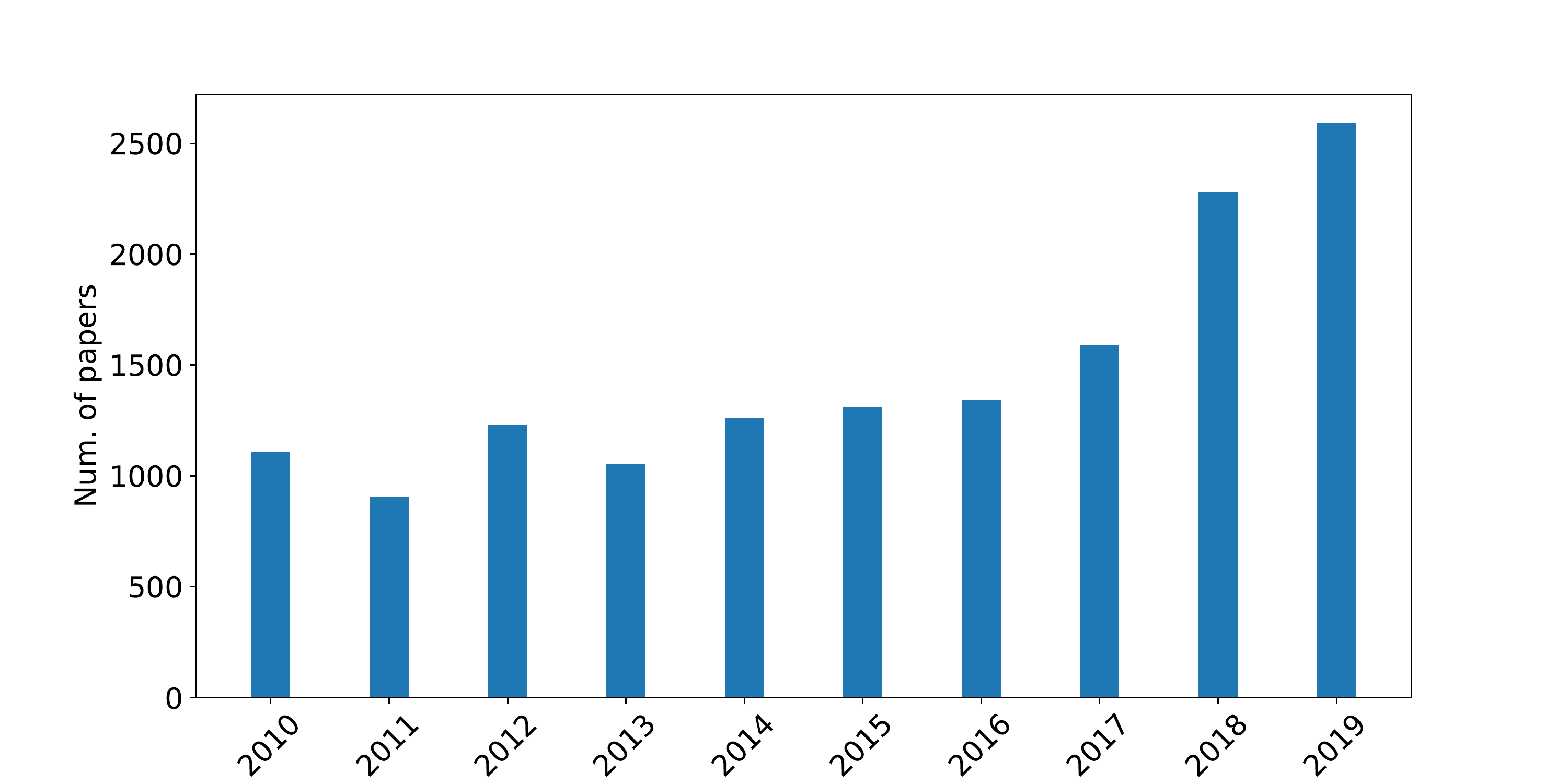}
  \caption{Total number of published papers.}
  \Description{Number of papers published in major conferences.}
  \label{number_of_papers}
\end{figure}

\begin{figure}[t!]
  \centering
  \includegraphics[width=0.85\linewidth]{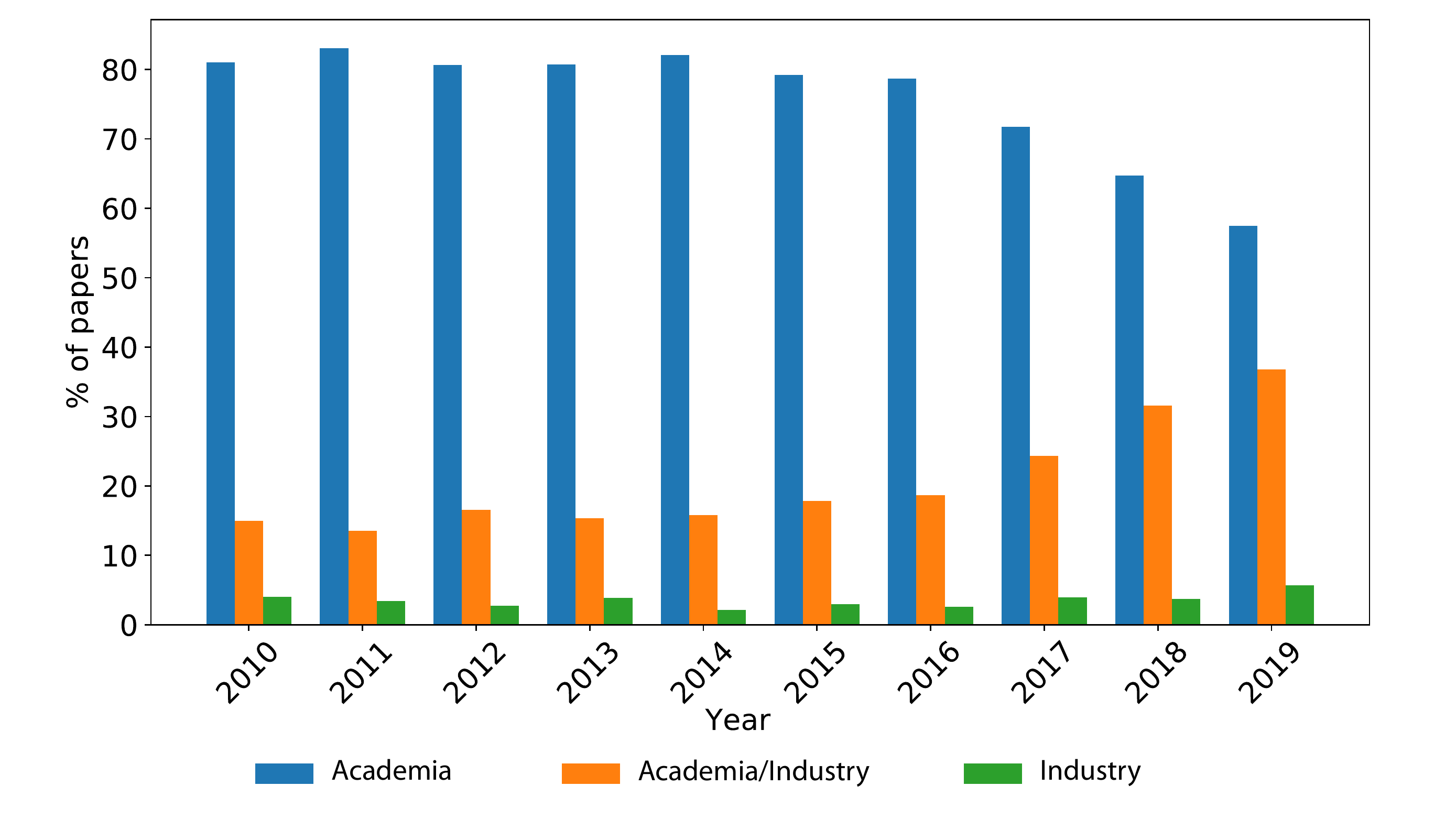}
  \caption{Percentage of papers from industry and academia.}
  \Description{Papers from industry/academia published in major conferences.}
  \label{industry_papers}
    \vspace{-1.5em}
\end{figure}

\begin{figure}[t!]
  \centering
  \includegraphics[width=0.85\linewidth]{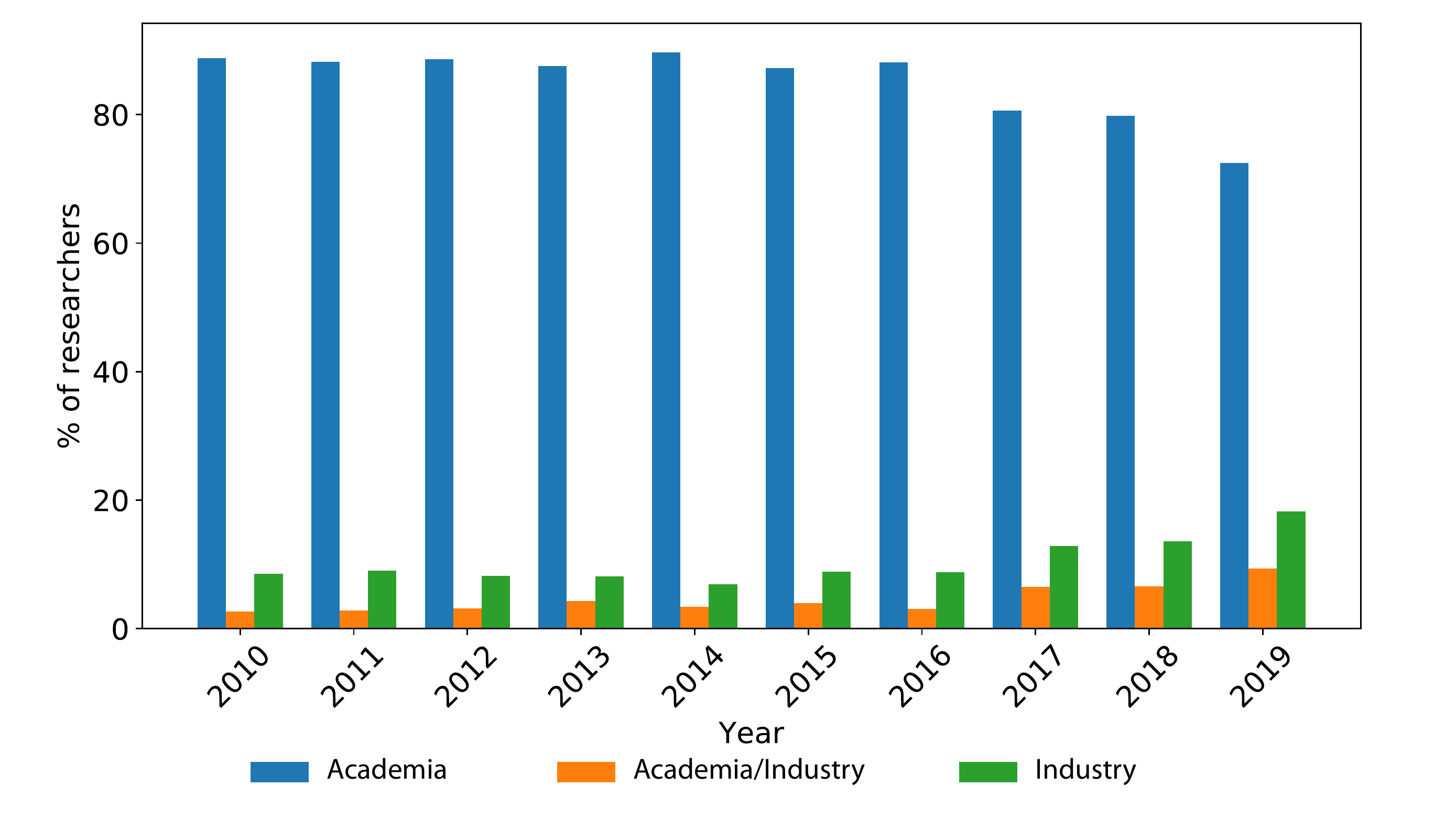}
  \caption{Percentage of authors affiliated with industry and/or academia.}
  \Description{Researchers from industry/academia.}
  \label{industry_researchers}
    \vspace{-1.5em}
\end{figure}

\begin{figure*}[t!]
  \centering
  \includegraphics[width=0.8\linewidth]{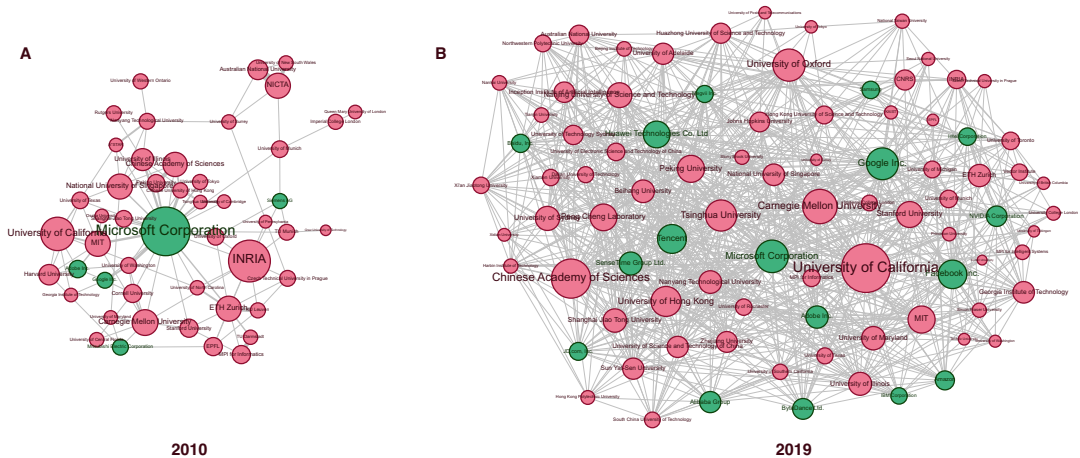}
  \caption{Part of the affiliation network in 2010 (A) and 2019 (B) showing academia (red nodes) and industry (green nodes) collaborations. Edges and nodes are weighted by the number of papers to which the affiliations contributed. In both networks, only nodes with node degree within top-10\% are shown for clarity.}
  \vspace{-1.5em}
  \label{affiliation_network}
\end{figure*}

\begin{figure}[t!]
  \centering
  \includegraphics[width=\linewidth]{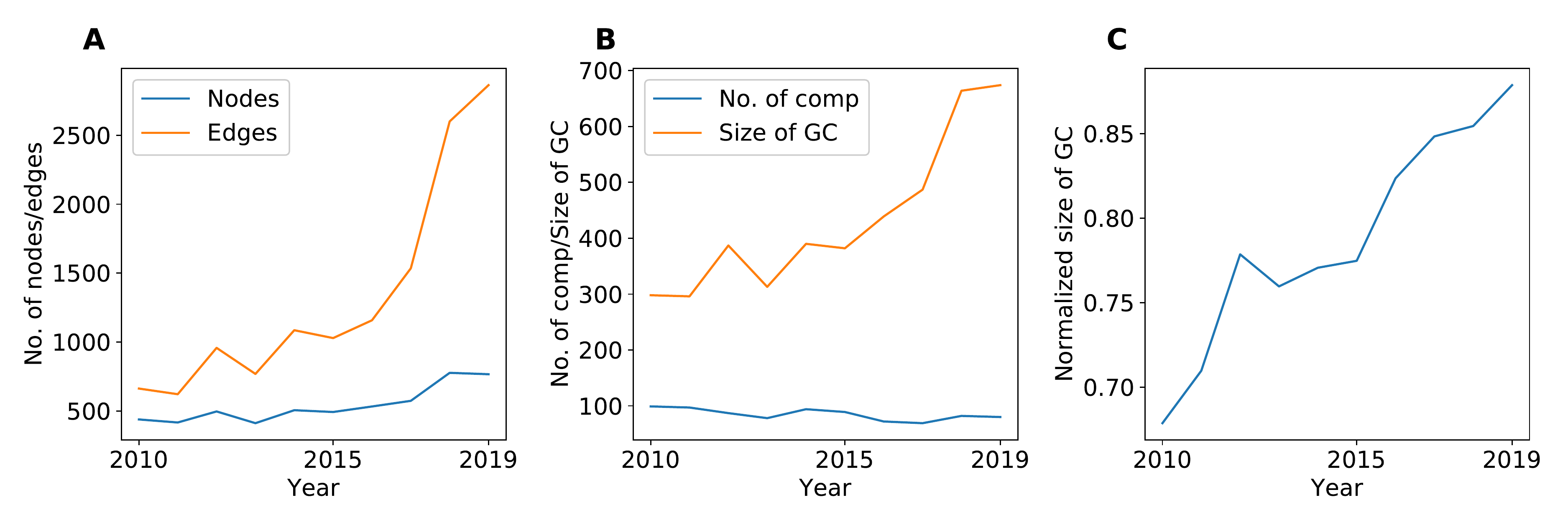}
  \caption{Affiliation network properties. A) Number of nodes and edges in affiliation graph for each year. B) Number of components and the size of the giant connected component for each year. C) Normalized size of the giant connected component (divided by the total number of nodes in the network) for each year.}
\vspace{-1.5em}
  \label{affiliation_network_properties}
\end{figure}

\textbf{Networks.} Besides computing descriptive statistics for the dataset, we also used the data to construct several networks: 1) affiliation network (representing collaborations on the organization level), 2) co-authorship network, and 3) citation network.

In the \textit{affiliation network}, nodes represent individual organizations. Each node's weight is incremented for every paper where at least one author was involved from the corresponding affiliation. An edge is added between two nodes if the authors from the corresponding affiliations co-authored a paper. The edge weight is incremented for every paper shared between these affiliations. Affiliation type, academia or industry, is assigned to every node. A separate network is constructed for each year of observation. 

The \textit{co-authorship network} is built in a similar fashion. A node is added for each author and its weight is incremented for every paper authored. Edges are created between the nodes representing co-authors and the weights are incremented for every paper co-authored. Each node is assigned an affiliation type: academia, if the author only published from academic institutions, or industry, if he/she published any papers with companies. A separate network is constructed for each year of observation. 

Finally, the \textit{citation network} contains nodes for each paper and edges between papers that cite each other. As before, nodes representing papers are assigned with the corresponding affiliation attribute (industry or academia).

\section{Growing industry participation}

\textbf{Contributions from academic and industry research.} There has been considerable growth in publications in the field of computer vision during the past decade. For instance, the number of publications in the top computer vision conferences has been steadily rising at an average annual growth rate of $12\%$ per year from approx. 1,000 papers published in 2010 to nearly 2,500 papers in 2019 (Figure \ref{number_of_papers}). To establish the industry contribution to this growth, we first looked at the percentage of the papers from the industry and academia published in the top vision conferences. As shown in Figure \ref{industry_papers}, the overall portion of papers where all or some researchers were from the industry increased in the span of 10 years. The number of papers published only by the industry-affiliated authors fluctuated between $2\%$ and $6\%$, however, the number of papers with authors both from the academia and industry more than doubled from $15\%$ to $37\%$. Together, papers with some industry involvement comprise over $40\%$ of all publications in 2019 compared to less than $20\%$ a decade earlier. 

\textbf{Changes in affiliations.} Similarly, the number of researchers affiliated with the industry has been growing steadily (Figure \ref{industry_researchers}). In particular, the percentage of authors affiliated only with companies grew from $7\%$ in 2010 to $17\%$ in 2019. The number of authors with mixed industry/academia affiliations also increased from $<1\%$ to $9\%$ in the same period.

%\begin{figure}[t!]
%  \centering
%  \includegraphics[width=\linewidth]{perc_top_100_industry.eps}
%  \caption{Percentage of top-100 authors affiliated with industry.}
%  \label{top_100_researcher_affiliations}
%\end{figure}

Most important is the behavior of the influential researchers in the field, defined as the top-100 authors with the highest PageRank score in the co-authorship network. In the observed period, the proportion of top researchers affiliated with the industry increased from $20\%$ in 2010 to nearly $60\%$ in 2019.

\textbf{Collaborations between industry and academia.} On both author and affiliation levels, there has been a trend towards more collaborations. The number of authors per paper has been gradually increasing. In academia, there were on average $3$ authors per publication in 2010 and $4.5$ in 2019. Papers with mixed academia/industry involvement had 6 authors on average in 2019. The maximum number of authors on a single paper rose from $15$ in 2010 to $28$ in 2019. As Figure \ref{industry_papers} shows, most of these new collaborations are between industry and academia.

The affiliation network helps show the trends in collaborations between the institutions and regional patterns. The affiliation network for 2010 was more disconnected, with the giant connected component covering $60\%$ of the $439$ nodes. In 2019, the total number of affiliations increased to $767$ and the giant component included $88\%$ of the nodes (Figure \ref{affiliation_network_properties}). In particular, the number of companies publishing their research increased significantly from $20$ in 2010 to over $200$ in 2019. In other words, industry affiliations comprised roughly one-quarter of all affiliations and contributed to $>40\%$ of all published papers last year.

We also investigated the important nodes in the affiliation network over the years of observation. In 2010, only one company, Microsoft, was among the top-10 affiliations (based on PageRank score), the rest were academic institutions. A decade later, four corporations, Google, Facebook, SenseTime, and Tencent, were among the top-10 contributors. Most of the top industry affiliations are fast-growing tech companies and start-ups that heavily rely on computer vision and machine learning for achieving their business goals, e.g. marketing, social networks, and autonomous driving.
 
Connections between academic and industrial affiliations also show regional patterns (see Figure \ref{affiliation_network}). For instance, researchers from Google and Facebook often collaborate with top US universities such as CMU and the University of California. Microsoft research divisions in Asia and Europe also have strong connections with major schools in those regions (e.g. ETH Zurich in Switzerland and Tsinghua University in China). A number of Chinese corporations, such as Huawei, Tencent, and SenseTime, collaborate with the largest Chinese and Australian universities.

\balance
\section{Impact of the industry presence}
\label{industry_effect}
The previous section confirmed growing industry presence in terms of the number of publications, changes in author affiliations, and a growing number of companies contributing to the research in the field. However, measuring the impact of the industry is much more difficult to accomplish since the industry presence could affect the field of computer vision in multiple ways that are difficult to quantify. Arguably, companies can offer better conditions, resources, and monetary compensation for their employees than most academic institutions. As a result, both recent graduates and established researchers may shift their research interests towards topics that will increase the likelihood of securing a job in the industry. Some corporations also offer various grants to researchers and establish research labs within institutions. Presumably, such grants and collaborations often go to researchers whose work is of potential interest to the companies. However, such indirect influences are not necessarily disclosed and thus are difficult to quantify.

To minimize subjective judgment, we will look at two areas where industry presence could have a measurable effect: 1) the evolution of the research topic trends and 2) citation preferences over the past decade.

\subsection{Research topic trends}

\begin{table}[t!]
\centering
\resizebox{\columnwidth}{!}{%
\begin{tabular}{l|l|l}
\multicolumn{1}{c|}{\#} & \multicolumn{1}{c|}{Topic} & \multicolumn{1}{c}{Keywords}                        \\ \hline
0                       & Text captioning            & generate text caption generation network            \\
1                       & Clustering                 & cluster datum space kernel representation           \\
2                       & Deep convolutional networks              & network deep feature neural convolutional           \\
3                       & Face recognition           & face recognition facial dictionary age              \\
4                       & Pose estimation            & pose estimation human part body                     \\
5                       & Super-resolution           & high resolution quality low filter                  \\
6                       & Action recognition         & video action temporal frame sequence                \\
7                       & Metric analysis            & metric similarity rank distance user                \\
8                       & 3D object reconstruction   & shape surface reconstruction object deformation     \\
9                       & Object tracking            & track tracking target object appearance             \\
10                      & Depth estimation           & depth camera scene motion estimation                \\
11                      & Point set registration     & point line solution robust set                      \\
12                      & Supervised learning        & training class datum label domain                   \\
13                      & Multi-view matching        & view match feature person matching                  \\
14                      & Constrained optimization   & optimization function registration solve constraint \\
15                      & Local feature descriptor   & local descriptor patch color texture                \\
16                      & Optimization               & time accuracy fast large efficient                  \\
17                      & Object recognition		 & visual object feature recognition attribute         \\
18                      & Semantic segmentation      & segmentation graph structure semantic label         \\
19                      & Object instance segmentation & object detection instance segmentation region       \\ \hline
\end{tabular}
}%
\caption{20 topics identified via LDA analysis.}
\vspace{-1.5em}
\label{lda_topics}
\end{table}

Latent Dirichlet Allocation (LDA) \cite{blei2003latent} is a commonly used technique for topic discovery in text data. The text corpus for this problem consists of titles and abstracts of the 14,686 papers. Following the typical pipeline, text data is pre-processed (removing stopwords, performing stemming, and lemmatization) before applying LDA. We used LDA implementation provided with the \texttt{gensim}, a Python framework for topic modeling. Parameters $alpha$ and $beta$ of the model were set to $0.01$ and $0.1$ respectively. The number of topics varied from $10$ to $200$ with a step of $5$. However, it appears that beyond $40$ topics the model stopped producing coherent results. We further investigated models with topics between $10$ and $40$ and found $20$ to be the optimal set. Additional hyper-parameter search within this interval did not improve over the baseline.

A common challenge with LDA is identifying an appropriate number of topics for the dataset. The topic coherence technique \cite{roder2015exploring} measures the degree of semantic similarity between the high-scoring words in the topic and helps select topics that are easier to interpret. In addition to coherence, we also looked at top-5 papers for each topic to verify the correctness of the results. Another useful tool was \texttt{pyLDAvis} for the interactive visualization of topics discovered by LDA. 

Since topics generated by LDA are sets of words, an additional step is needed to label the topics using domain knowledge. The final set of $20$ topics, corresponding top-5 words, and topic labels are shown in Table \ref{lda_topics}.

\begin{figure}[t!]
 \centering
 \includegraphics[width=\linewidth]{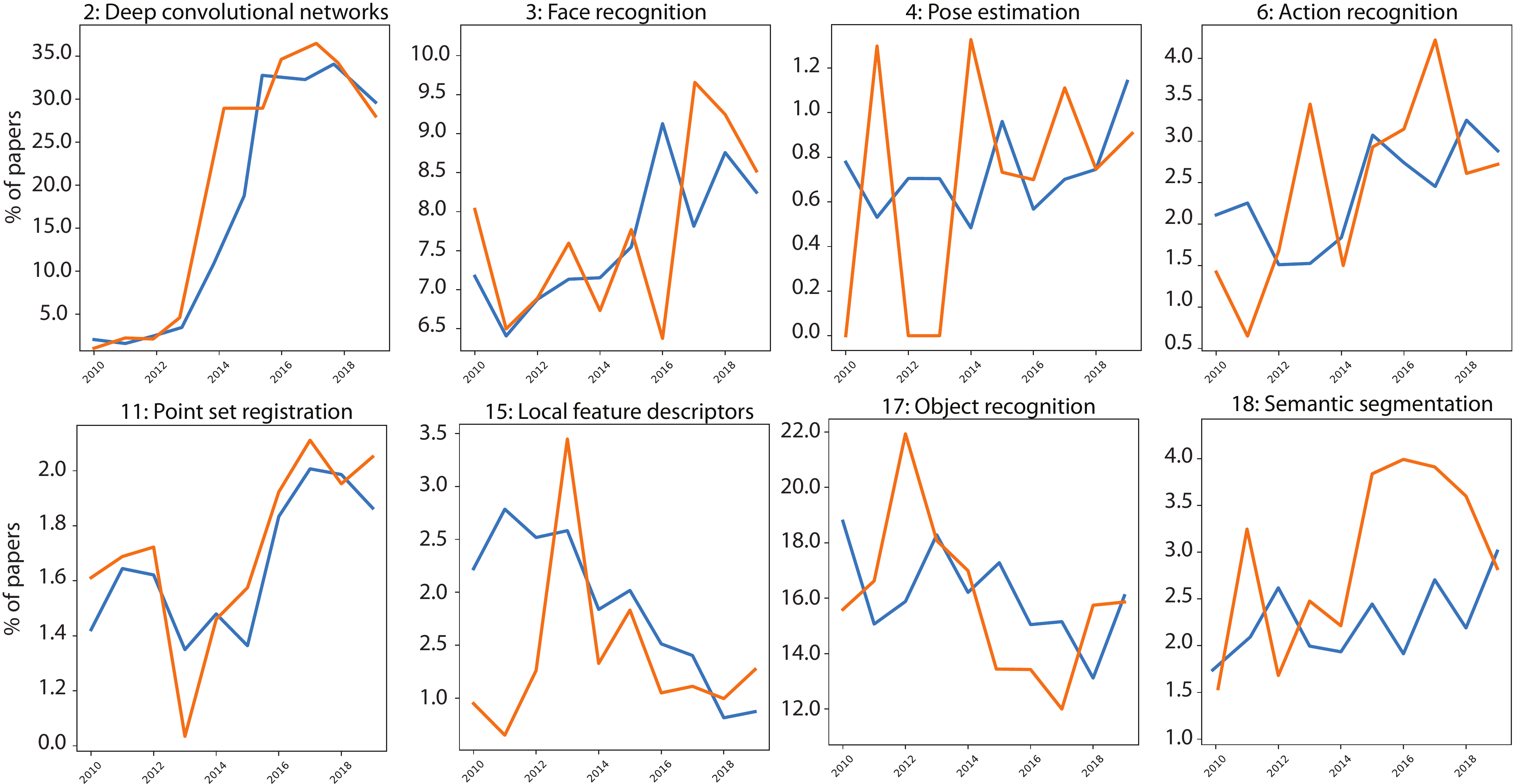}
 \caption{Plots of normalized paper counts for select topics. Blue and yellow lines show academic and industry paper counts respectively.}
 \label{auto_20_selected}
\end{figure}

To identify topic trends, we used the trained LDA model to find the top topic for each document in the dataset and then aggregated counts for each topic per year, separately for academia and industry. These counts are then normalized by the total number of publications for that year in academia and industry, respectively. Select plots of normalized paper counts are shown in Figure \ref{auto_20_selected}. For each plot, the topic index and label are shown.

Overall, trends in research topics were evolving similarly in academia and industry. To verify this, we also plotted topic trends using LDA models with different numbers of topics which also did not show significant differences. Another confirmation of the plot correctness is that the overall trends align with the known trends in computer vision research. For instance, there has been growing interest in deep convolutional networks, the number of publications on face recognition, pose estimation, action recognition, and point set registration are also growing due to a large number of potential applications. Hand-crafted local feature descriptors were an important topic before the advent of neural networks, and the number of publications is on the decline. Similarly, object recognition was one of the first successful applications of deep learning and was heavily studied in the 2012-2014. As the benchmark results began to saturate, the number of papers dedicated to this topic decreased.

\subsection{Influence of industry on citation preferences}

A citation is, by definition, an explicit declaration of influence by other papers, therefore, citation network is a reasonable source of data for inferring possible influences on researchers' choices. It should be noted that authors cite only a fraction of the papers that influenced them, and citation, in general, is subject to multiple biases, such as preference towards citing papers with already high citation count, papers with multiple co-authors, and publicly available implementations/data \cite{macroberts2010problems, onodera2015factors, vandewalle2019code}. In this section, we will investigate whether papers produced by the industry have a larger influence on the community, i.e. are more likely to be referenced from the pool of other similar articles.

The citation network built for this part of the analysis is a directed graph with 14,686 nodes and 161,514 edges. Only citations of the papers within the top-5 conferences were considered out of the total 600K citations parsed from the paper bibliographies. More than 95\% of the network's nodes are within the giant connected component, and nearly 25\% of papers are uncited.

According to the data from the citation network, industry papers are cited much more frequently. Out of the top-10 cited papers, 2 are from the industry, 3 are of mixed academic/industry authorship, and 5 are from the academic authors only. These highly-cited papers introduce widely used architectures (ResNet \cite{he2016deep}), algorithms for common vision tasks (e.g. object detection \cite{girshick2015fast, liu2016ssd}) and commonly used datasets (e.g. COCO \cite{lin2014microsoft} for object detection and segmentation and KITTI \cite{geiger2012we} for autonomous driving). On average, industry papers were cited $23$ times vs $11$ for academia/industry and $10$ for academic papers. Note also that in the citation network academic papers comprise $73\%$, while only $4\%$ of the papers were contributed by the industry and $23\%$ by mixed academia/industry groups of authors. Hence, the $4\%$ of industry papers gathered $8\%$ of all citations.

\begin{figure}[t!]
 \centering
 \includegraphics[width=0.85\linewidth]{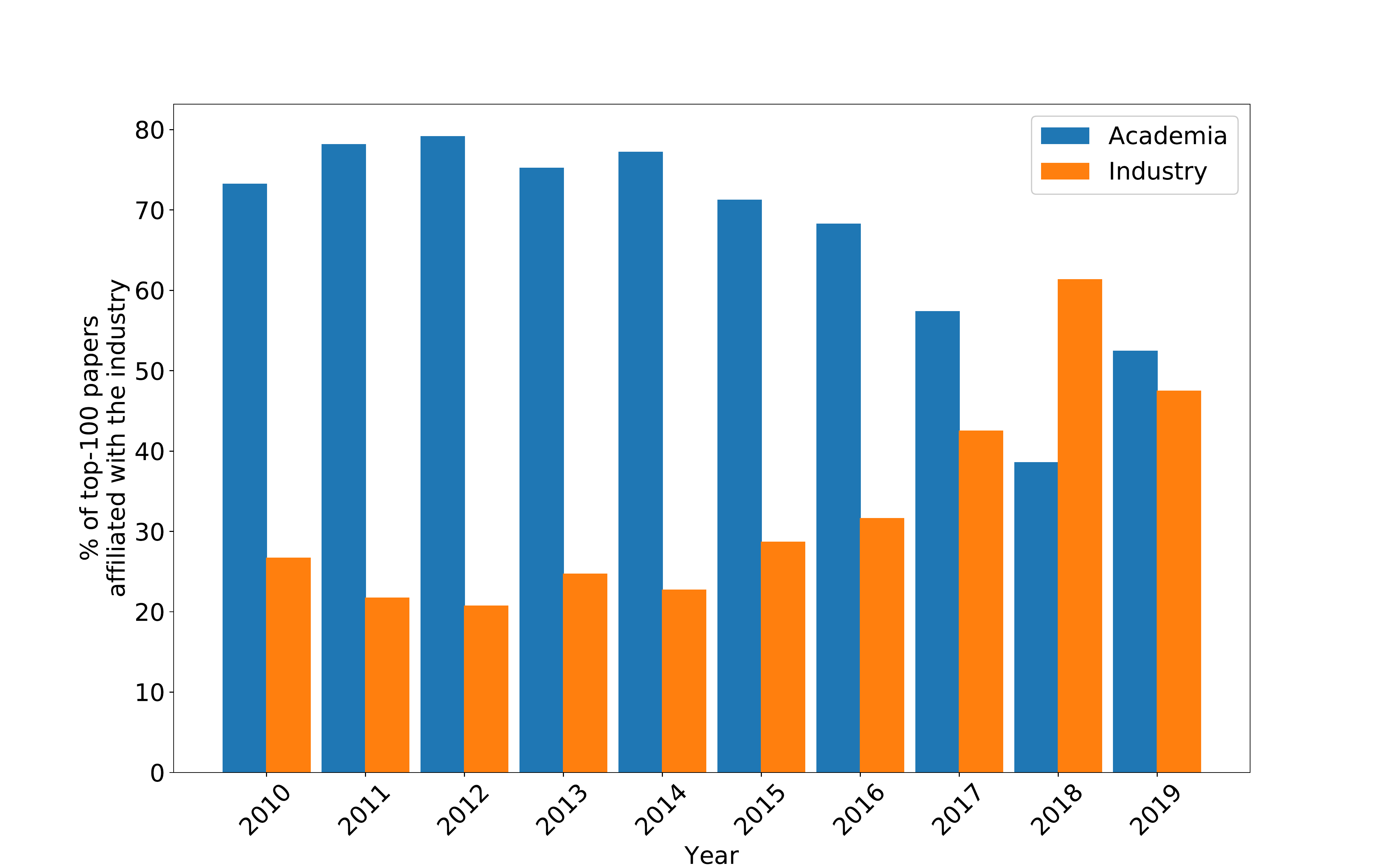}
 \caption{Proportion of academia and industry papers within top-100 papers (with highest PageRank score) for each year.}
 \label{perc_top_100_papers}
\end{figure}

A different way of assessing the influence of papers in the citation network is by looking at their PageRank score. Figure \ref{perc_top_100_papers} shows the proportion of academia and industry papers within top-100 papers with the highest PageRank. It can be seen that since 2015, industry-affiliated papers increased their influence, briefly overtook academic papers in 2018 and are nearly equal in influence to academic papers in 2019. Nevertheless, $67$ out of $100$ top papers over the whole decade were produced by academia. 

\begin{figure}[t!]
 \centering
 \includegraphics[width=0.85\linewidth]{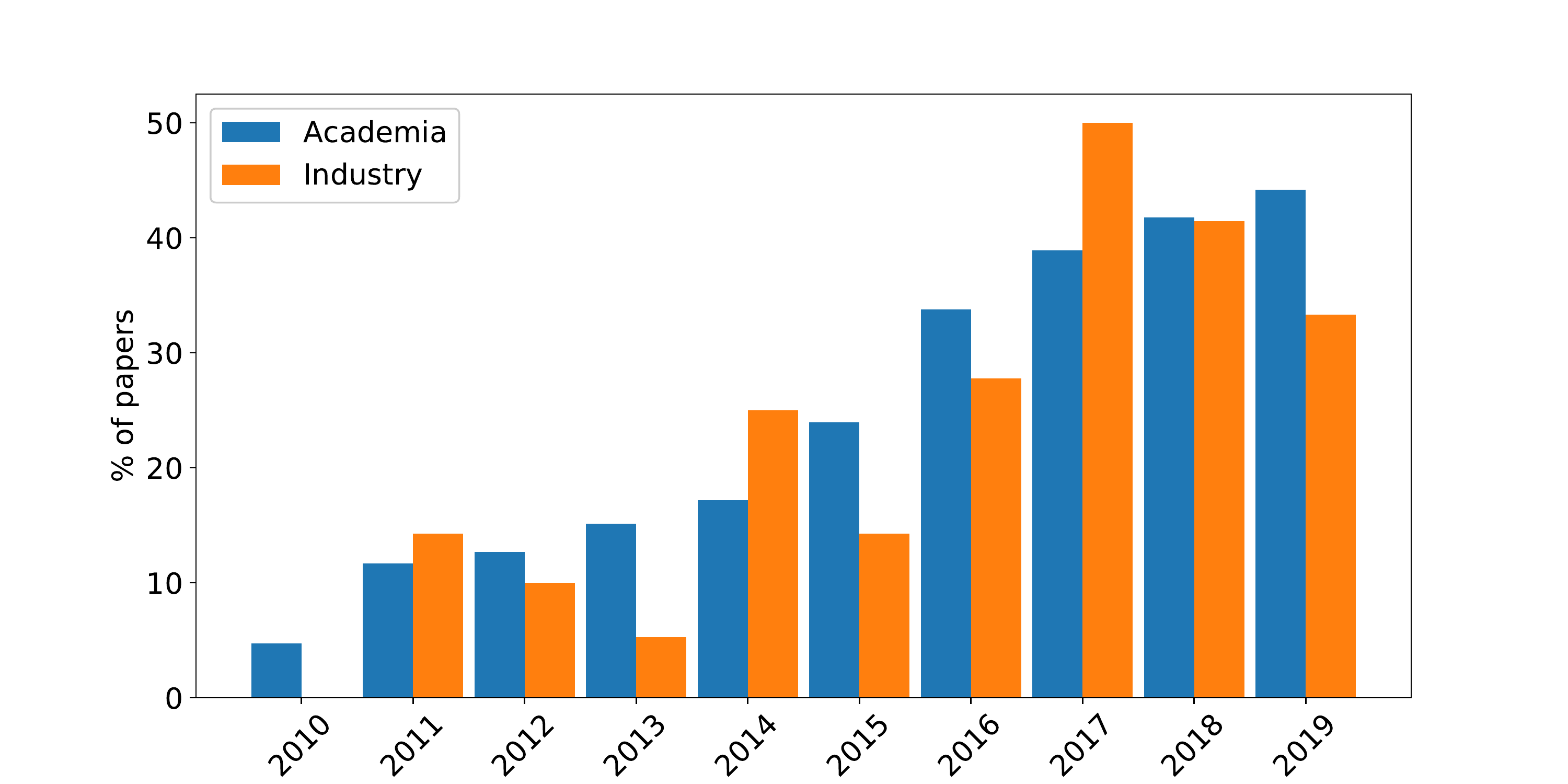}
 \caption{Proportion of academia and industry papers with publicly available code per year.}
 \label{perc_papers_with_code}
\end{figure}

To further investigate the citation bias towards industry papers, we looked at the availability of code/data associated with the papers since it has been shown to give significant citation advantages in other areas of science \cite{vandewalle2019code}. The relative proportion of papers with code has been growing steadily as shown in Figure \ref{perc_papers_with_code}. The proportion of publicly released code and the growth rate of the number of papers with code was approximately the same regardless of whether industry-affiliated authors were involved. Therefore this factor can be ruled out as the reason for higher average citations of industry papers.

Another potential aspect of citation bias is the influence of other papers. Typically, when researching a topic, references from other similar papers are a common source of expanding bibliography. Therefore, the aim of this analysis is to verify whether researchers cite industry papers because they came across other papers that do so.

In general, the problem of distinguishing co-occurrence from the influence has been addressed in other examples of real networks, such as social networks. In particular, one of the ways to identify the presence of influence in a network is a \textit{shuffle test}, first proposed in \cite{anagnostopoulos2008influence}. This test assumes that correlation is time-invariant and probes whether there is a temporal dependency in the activation of the node and activations of the node's friends in the previous time step. A variant of this method proposed in \cite{papagelis2011individual} will be applied to this problem.

Following \cite{papagelis2011individual}, in the citation graph $G(V, E)$ where vertices represent papers and edges correspond to citations there could be active and inactive nodes. When the paper in the dataset cites a paper from the industry, the corresponding nodes will be considered active. At the time $t=0$ all nodes are inactive, at $t=T$, some nodes in the network will be activated. Then, an ordered set of nodes $A={v_1, v_2, ..., v_k}$ corresponds to papers that cited industry-associated papers during the observed period $[0,T]$, where $v_i$ became active at time $t_i$. 

To find industry influence, another set of active nodes $A_{II}$ is constructed. It contains active nodes at least one of whose neighbors also cited an industry paper. The industry influence can be represented by the ratio between the two sets given by $II=\frac{A_{II}}{A}$. 

To test whether node activations due to influence are random, a copy of the graph \textit{G} and a set of activations $A$ is created. Except, in this copy of the graph, the times when the nodes became activated are randomly permuted. The activations due to neighbor influence are then re-established in this new instance of a problem. The industry influence here is is given by $II_{shuffled}=\frac{{A'}_{II}}{A}$.

$II$ and $II_{shuffled}$ values can be compared for every time step in the observed period. If the original $II$ values are higher than $II_{shuffled}$ at every timestep and for a different number of activated neighbors, then it can be concluded that citing industry papers is likely due to the industry influence rather than the similarity between the papers.

\begin{figure}[t]
 \centering
 \includegraphics[width=0.48\linewidth]{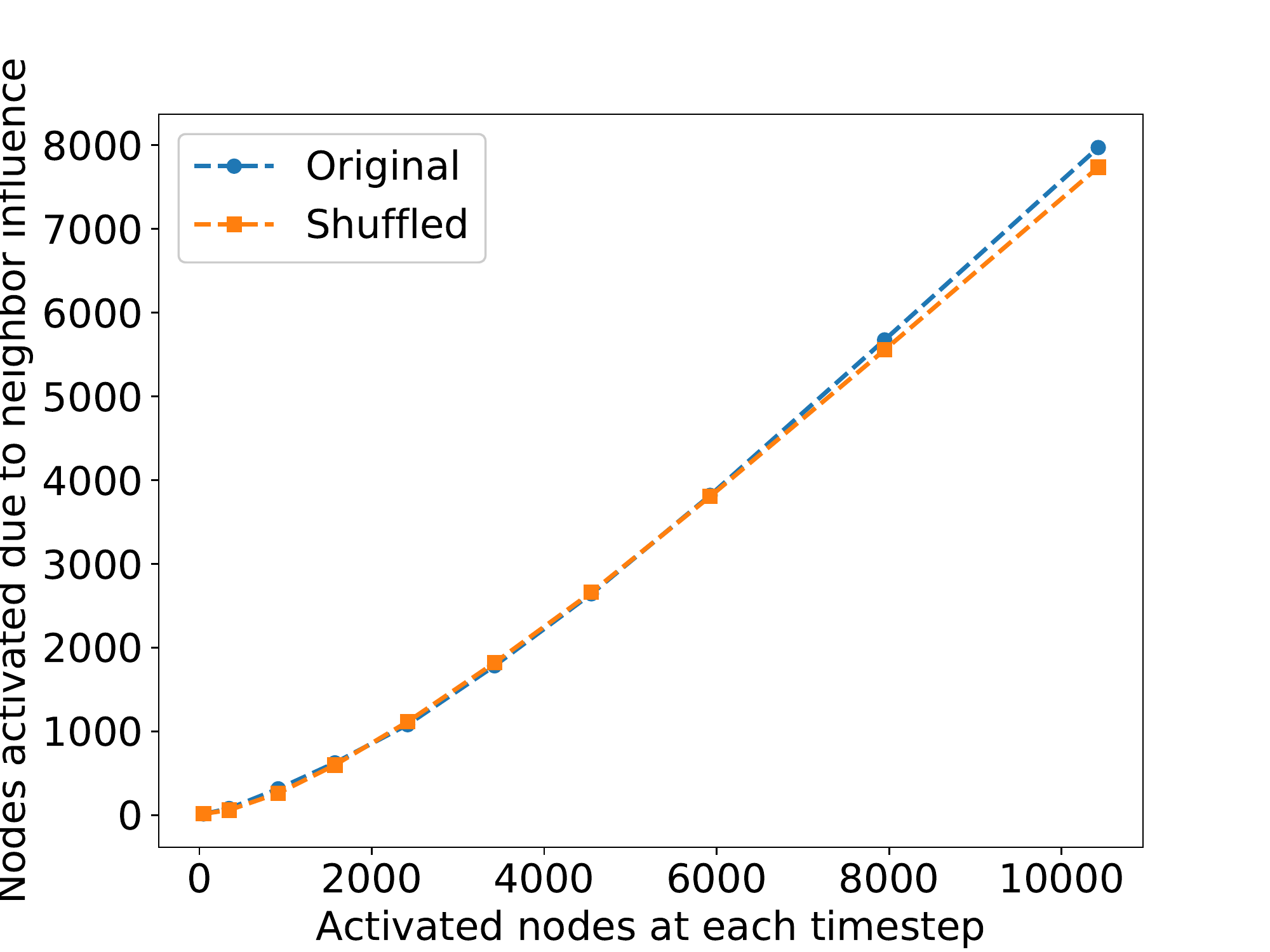}
 %\caption{Plot of the original vs shuffled set of activations.}
 \includegraphics[width=0.48\linewidth]{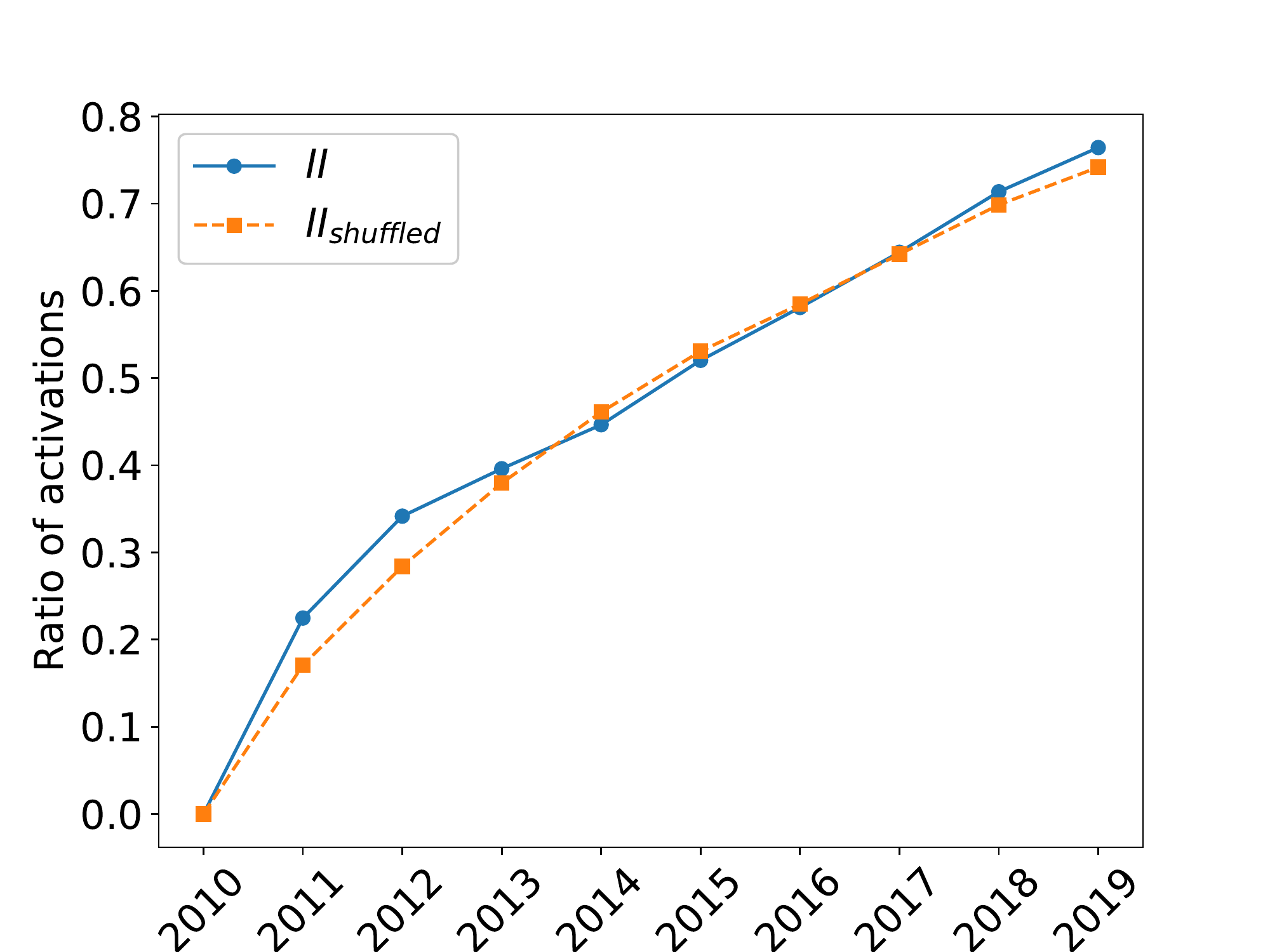}
 %\caption{Plot of the original vs shuffled set of activations.}
 \caption{Plot of the original vs shuffled set of actiovations (left) and ratios (right).}
 \label{shuffle_test}
\end{figure}

The plots in Figure \ref{shuffle_test} show the absolute and relative sizes of activation sets in the original and shuffled conditions. Since the time steps are permuted randomly, the results are averaged over $100$ runs of the test. Note that between 2012 and 2014 and after 2017 $II$ ratios in the original condition become slightly higher than in the shuffled condition showing some influence towards citing industry papers. But, overall, it cannot be concluded that the presence of industry citations in other papers plays a significant role in boosting the number of industry paper citations.

\section{Conclusions}
The goal of this study was to investigate what is the share of industry-sponsored research in the field of computer vision and how such industry presence affects the development of the field. The analysis was based on nearly 15K papers published between 2010 and 2019 in the top-5 computer vision conferences.

The first part of this study demonstrated an increasing growth of industry presence in the computer vision community. There are now more than $200$ actively publishing companies (including corporations and start-ups). While the proportion of papers authored exclusively by the industry-affiliated researchers remains small, since 2016, an increased number of collaborations between companies and academic institutions is observed. The publications resulting from these collaborations account for nearly half of all papers accepted to top-5 conferences in 2019.

The second part attempted to quantify the effect of industry presence on two aspects of research: research topic trends and citation preferences. Although topic trends were shown to be similar across academia and industry, there appears to be a strong preference for citing industry papers, which may be interpreted as them having higher importance. To further analyze possible causes for such citation bias, two possible causes were considered: the availability of code/data and influence from other papers in the citation network. The first cause was ruled out since both academic and industry publications provided code in the observed period with equal (and increasing) frequency. There was evidence of some influence from other papers citing industry publications, particularly in 2012-2014 and after 2017. However, this factor alone cannot justify a much higher average number of industry paper citations.

There could be several reasons for such results. Since there were too few purely industry-affiliated papers, all analysis was based on the papers with mixed affiliations. This is potentially problematic, since these publications may not properly reflect the state of research in the industry. Given that many industry-affiliated researchers are also part of the peer-review process, there could be an additional bias as to what publications are accepted or cited. Lastly, the industry involvement in computer vision research started increasing relatively recently, around 2016. Perhaps, not enough time has passed, and another 5-10 years will be needed to see the effects of industry involvement.

%%
%% The next two lines define the bibliography style to be used, and
%% the bibliography file.
\bibliographystyle{ACM-Reference-Format}
\bibliography{acmart}

\end{document}